# Introduction of the weight edition errors in the Levenshtein distance

Gueddah Hicham
Telecom and Embedded Systems Team, SIME Lab ENSIAS, University of Mohammed V Souissi
Rabat, Morocco

Yousfi Abdallah
Faculty of juridical, Economic and Social Sciences
University Mohammed V Souissi
Rabat, Morocco

Belkasmi Mustapha
Telecom and Embedded Systems Team, SIME Lab ENSIAS, University of Mohammed V Souissi
Rabat, Morocco

*Abstract*—**In this paper, we present a new approach dedicated to correcting the spelling errors of the Arabic language. This approach corrects typographical errors like inserting, deleting, and permutation. Our method is inspired from the Levenshtein algorithm, and allows a finer and better scheduling than Levenshtein. The results obtained are very satisfactory and encouraging, which shows the interest of our new approach.**

**Keywords-component; spelling errors; correction; Levenshtein distance; weight edition error.**

## I. Introduction

Automatic correction of spelling errors is one of the most important areas of natural language processing. Research in this area started in the 60s [1]. Spell checking is to find the word closest to the erroneous word and words in the lexicon. This approach is based on the similarity and the distance between words.

In the areas we are interested in the treatment of misspelling out of context; several studies have been achieved to present methods of automatic corrections. Among these works, we cite:

- The first studies have been devoted to determining the different type of elementary spelling error, called editing operations [2] which are:
  - Insertion: add a character.
  - Deletion: omission of a character.
  - Permutation: change of position between characters.
  - Replacement: replace a character with another.
- Based on the work of Damerau, Levenshtein [3] considered only three editing operations (insertion, deletion, permutation) and defined his method as edit distance. This distance compares two words by calculating the number of editing operations that transforms the wrong word to the correct word. This distance is called Damerau-Levenshtein distance.
- Oflazer [4] proposed a new approach called "Error tolerant Recognition", based on the use of a dictionary represented as finite state automata. According to this approach, the correction of an erroneous word is to browse an automata-dictionary for each transition by calculating a distance called cut-off edit distance, and stack all the transitions not exceeding a maximum threshold of errors. Savary [5] proposed a variant of this method by excluding the use of cut-off edit distance.
- Pollock and Zamora [6] have defined another way to represent a spelling error by calculating the so called alpha-code (skeleton Key), hence the need for two dictionaries: a dictionary of words and other for alpha-codes. Therefore, to correct an erroneous word, we extract its alpha-code and comparing it to the alpha codes closest. This method is effective in the case of permutation errors. Ndiaye and Faltin [7] proposed an alternative method of alpha code, who defined a system of suitable spelling correction for learning the French language, based on the method of alpha code modified by combining other techniques such as phonetic reinterpretation, in case where the first method does not find solutions.
- A critical analysis of existing systems for spell checker, realized by Souque [8] and Mitton [9], confirms that these systems have limitations in the proposed solutions to some type of erroneous word.

In the work presented in this paper, we propose a new metric approach inspired from the Levenshtein algorithm. This approach associates for each comparison between two words a weight, which is a decimal number and not an integer. This weight allows the better and perfect scheduling solutions proposed by the correcting system of the spelling errors.

## II. Levenshtein algorithm

The metric method developed by Levenshtein [3], measures the minimal number of elementary editing operations to transform one word to another. The minimum term was defined by Wagner and Fischer [1] thus proposing the programming dynamic technique to solve the edit distance. Elementary editing operations considered by Levenshtein are:

- Insertion: Add a character ' ' (    )
- Deletion: omission of the character ' ' (    )
- Permutation: replacement of the character ' ' with a ' ' (    )

The calculation procedure of the Levenshtein distance between two strings $X = x_1x_2...x_m$ of length m and $Y = y_1y_2...y_n$ of length n, consists in calculating recursively the edit distance between different substrings of $X$ and $Y$.
The edit distance between the substrings $X_1^i = x_1x_2...x_i$ and $Y_1^j = y_1y_2...y_j$ is given by the following recursive relationship:

$$D(i,j) = D(X_1^i, Y_1^j)$$
$$D(i,j) = \min\{D(i-1,j)+1, D(i,j-1)+1, D(i-1,j-1)+cost\}$$

With $cost = \begin{cases} 0 & if\ x_{i-1} = y_{j-1} \\ 1 & else \end{cases}$

and the following initializations: $D(i, \varepsilon)=i$ and $D(\varepsilon, j)=j$, where $\varepsilon$ represents the empty string.

**Example**:

TABLE I. CALCULATION OF EDIT DISTANCE

|   |   |   |   |   |   |   |
|---|---|---|---|---|---|---|
|   | 0 | 1 | 2 | 3 | 4 | 5 |
|   | 1 | 0 | 1 | 2 | 3 | 4 |
|   | 2 | 1 | 0 | 1 | 2 | 3 |
|   | 3 | 2 | 1 | 1 | 1 | 2 |
|   | 4 | 3 | 2 | 1 | 2 | 2 |
|   | 5 | 4 | 3 | 2 | 2 | 2 |

The matrix shows the recursive calculation of the Levenshtein distance between the erroneous word " " and the dictionary word " ", the distance is 2.

The limitation of such a spelling correction system using the edit distance is not to allow a correct order of suggested solutions to a set of candidates having the same edit distance.
For example, we have the dictionary word "السيف" and the erroneous word "السيق", the Levenshtein method returns the same edit distance for the following set of words.

| Erroneous word | Dictionary words | Edit distance |
|---|---|---|
| السيق |   | 1 |
|   |   | 1 |
|   |   | 1 |
|   | السيف | 1 |
|   | السين | 1 |
|   | الشيق | 1 |

In order to remedy to this limitation, we propose an adaptation of the Levenshtein distance. This adaptation gives a better scheduling of the solutions having the same edit distance.

### III. LEVENSHTEIN METHOD ADJUSTED

To remedy to the scheduling problem, we introduced the frequency of the three type errors of the editing operations.
We carried a test with four experienced users: they have typed a set of Arabic documents in order to calculate the frequency error of the editing operations. For this, we define the following three matrices:
- Matrix frequency of insertion error.
- Matrix frequency of deletion error.
- Matrix frequency of permutation error.

In this context, we modified the Levenshtein distance between two words by taking into account these three matrices.
More formally, for two strings $X = x_1x_2x_3..x_m$ of length m and $Y = y_1 y_2 y_3..y_n$ of length n, the calculation procedure of the measurement between X and Y is done in the same manner as that of Levenshtein algorithm, but introducing the matrices frequency of the editing errors. This measure $\mathcal{M}(i,j)$ is given by the recursive relationship:

$$\mathcal{M}(i,j) = \min\{\mathcal{M}(i-1,j)+1-\mathcal{F}_{aj}(x_{i-1}), \mathcal{M}(i,j-1)+1-\mathcal{F}_{sup}(y_{j-1}), \mathcal{M}(i-1,j-1)+cost\}$$

With $cost = \begin{cases} 0 & if\ x_{i-1} = y_{j-1} \\ 1 - \mathcal{F}_{permut}(x_{i-1}/y_{j-1}) & else \end{cases}$

And
- $\mathcal{F}_{aj}(x_i)$ = the error frequency of adding the character '$x_i$' in a word.
- $\mathcal{F}_{sup}(y_j)$ = the error frequency of deleting the character '$y_j$' in a word.
- $\mathcal{F}_{permut}(x_i/y_j)$ = the error frequency of the permutation '$x_i$' with the character '$y_j$'.

For the algorithm, we take the following initializations:
- $\mathcal{M}(0,0)=0$
- $\mathcal{M}(i,0)=\mathcal{M}(i-1,0)+\mathcal{F}_{aj}(x_{i-1})$
- $\mathcal{M}(0,j)=\mathcal{M}(0,j-1)+\mathcal{F}_{sup}(y_{j-1})$

**Example**:
The measure between two words "السيق" and "السيف" is calculated in the following matrix:

|   |   |   |   |   |   |   |
|---|---|---|---|---|---|---|
|   | 0 | 0,1785 | 0,2525 | 0,2525 | 0,3771 | 0,3771 |
|   | 0,1111 | 0 | 0,9259 | 1,1414 | 1,2525 | 1,2660 |
|   | 0,2657 | 0,8454 | 0 | 1,0000 | 1,8754 | 2,1114 |
|   | 0,2802 | 1,1017 | 0,9855 | 0 | 0,8754 | 1,8754 |
|   | 0,4058 | 1,2273 | 1,8599 | 0,8744 | 0 | 1,0000 |
|   | 0,4058 | 1,2273 | 2,1533 | 1,8744 | 1,0000 | 0,9911 |

The measure $\mathcal{M}$ (السيق, السيف) = 0,9911.

## IV. TESTS AND RESULTS

The statistical study that we have done is to determine the frequency of errors editing operations (insertion, deletion, permutation). For this, we launched a typing test of Arabic documents for a set of users.

Our training corpus is a set of Arabic documents typed by four expert users. From this corpus, we calculated the three matrices of error previously defined.

TABLE II.  STATISTIC ON EDITING ERROR

| Editing operation | Number of errors | Total |
|---|---|---|
| Insertion | 202 | |
| Deletion | 295 | 1420 |
| Permutation | 923 | |

To test our method, we have performed a comparison between our approach and that of Levenshtein for scheduling of the solutions. The implementation of our approach was performed by a developed program in Java language.

To compare our approach with that Levenshtein, we processed only 190 errors. The results obtained are summarized such as:

- Of 190 erroneous words, our method correctly classified 119 words in the first position, while the Levenshtein distance has only 19 classified in the first position. The rest of 119 was distributed on the 2nd, 3rd, 4th, 5th... 10th position. Statistically our method has proposed 62.63% of correct words in the first position against 10% for the Levenshtein.
- Of 71 erroneous words, our method has ranked in second position 40 solutions, while for these 40 erroneous words Levenshtein distance was only 15 classified in 2nd position and the remaining 25 were distributed on the 3rd, 4th, 5th, ..., 10th position with a rate of 21.05% against 7.89% for the Levenshtein.
- Of 30 erroneous words, our method proposed 21 corrections in the third position, while on these 21 words Levenshtein method proposed that 5 in 3rd position and the rest distributed over the posterior positions, giving a rate of 11.05% for our method and 2.63% for the Levenshtein distance.
- For the remaining 10 erroneous words, our method has positioned in the fourth positions whereas the edit distance has proposed the following: 3 in 4th, 3 in 5th, 2 in 8th and 2 in 10th with a rate of 5.26% for our method and 1.57% for the Levenshtein. The table below summarizes the results obtained.

TABLE III.  PERCCENTAGE OF SCHEDULING SOLUTIONS

| | First position | Second position | Third position | Fourth position |
|---|---|---|---|---|
| Lev. Meth. Adjusted | 62,63% | 21,05% | 11,05% | 5,26% |
| Levenshtein | 10% | 8% | 2,63% | 1,57% |

## V. CONCLUSION

In conclusion, we note the interest of our method in scheduling of correct words for the first and second positions while for the third and fourth positions can justify this by the unavailability of frequencies (zero frequency) for some Arabic alphabetic character during execution of our test.